\pdfoutput=1

\documentclass[11pt,table,dvipsnames]{article}

\usepackage{acl}
\usepackage{times}
\usepackage{latexsym}
\usepackage[T1]{fontenc}

\usepackage[utf8]{inputenc}

\usepackage{microtype}
\usepackage{latexsym}
\usepackage{dsfont}
\usepackage{booktabs} 
\usepackage{subfigure}
\usepackage{graphicx}
\usepackage{bigstrut}
\usepackage{multirow}
\usepackage{floatrow}

\usepackage[normalem]{ulem}
\usepackage{amssymb}
\usepackage{amsfonts}
\usepackage{multicol}
\usepackage{tipa}
\usepackage[ruled,linesnumbered]{algorithm2e}

\usepackage{paralist}
\usepackage[switch]{lineno}
\usepackage{multirow, makecell, caption}
\usepackage{verbatim}

\usepackage{csquotes}
\usepackage{xspace}
\usepackage{paralist}
\usepackage{mdwlist}
\usepackage{subfigure}
\usepackage{array}
\usepackage{enumitem}
\usepackage{bm}
\usepackage{colortbl}
\usepackage{dsfont}
\usepackage{url}
\usepackage{graphicx}
\usepackage{tabularx}
\usepackage{amsmath}
\usepackage{bbm}
\usepackage{tikz}
\usepackage{pgfplots}
\title{
Harnessing Knowledge and Reasoning for Human-Like \\Natural Language Generation: A Brief Review}

\author{
Jiangjie Chen\\
Fudan University\\
\texttt{jjchen19@fudan.edu.cn}\\
\And
Yanghua Xiao \\
Fudan University\\
\texttt{shawyh@fudan.edu.cn}
}

\begin{document}

\maketitle

\begin{abstract}
    The rapid development and application of natural language generation (NLG) techniques has revolutionized the field of automatic text production. 
    However, these techniques are still limited in their ability to produce human-like text that is truly reasonable and informative.
    In this paper, we explore the importance of NLG being guided by knowledge, in order to convey human-like reasoning through language generation. 
    We propose ten goals for intelligent NLG systems to pursue, and briefly review the achievement of NLG techniques guided by knowledge and reasoning. 
    We also conclude by envisioning future directions and challenges in the pursuit of these goals.
\end{abstract}

\section{Introduction}
\label{sec:intro}

Language, as the vehicle of thought~\cite{wittgenstein1958blue}, is one of the most fundamental means for humans to reason and communicate with each other.
Hence, the technology of natural language generation (NLG) has always been one of the main focuses throughout the history of AI research~\cite{weizenbaum1966eliza}.
In the age of modern deep learning, NLG techniques have been widely deployed in real-life applications, including machine translation~\cite{wu2016google}, news reporter~\cite{xu-etal-2020-xiaomingbot}, dialogue systems~\cite{thoppilan2022lamda}, automatic report generation~\cite{gkatzia-et-al-2017}, etc.
Beyond that, NLG models also serve as a rather universal workhorse in other types of NLP tasks, such as structured prediction~\cite{paolini2021structured}.

The achievement so far for NLG research has enabled the wide application of NLG techniques, but the dangers beneath them are still far from being resolved.
The foundation of NLG models has shifted over the years, from rule-based models~\cite{kukich-1983-design,mckeown1992text} to statistical models~\cite{vaswani2017attention}, and now at pre-trained language models (PLMs)~\cite{radford2019language,devlin-etal-2019-bert,raffel2020exploring,lewis-etal-2020-bart,NEURIPS2020_1457c0d6}.
However, rule-based models are too rigid to generate natural language.
Statistical models based on neural networks and training datasets suffer from limitations such as reporting bias, exposure bias, generalization issue, etc.
Recent years, PLMs have shown their excellent capabilities of understanding and generating natural language.
However, the research on PLMs does not necessarily solve the fundamental problems NLG, as the alleviation of these problems comes from exposing the models on much more data with self-supervised training.
Moreover, multiple problems still occur, including model explainability~\cite{liu-etal-2021-explainaboard}, hallucination~\cite{raunak-etal-2021-curious,xiao-wang-2021-hallucination}, ethical risks~\cite{vanderlyn-etal-2021-seemed,ziems-etal-2022-moral}, logical inconsistency~\cite{kim-etal-2020-will,elazar-etal-2021-measuring}, etc.

Towards these challenges, the research community has formulated an important perspective with symbolic knowledge, covering rules~\cite{Hwang2021COMETATOMIC2O}, commonsense knowledge~\cite{Speer_Chin_Havasi_2017}, world knowledge~\cite{10.5555/1785162.1785216,vrandevcic2014wikidata}, etc.
Since symbolic knowledge can be used to describe complicated concepts and their connections, it would be easier for the models to comprehend the textual world.
Therefore, symbolic knowledge brings significant opportunities towards NLG models in these respects: 
1) symbolic knowledge provides rich background materials for NLG models to generate from;
2) symbolic knowledge serves as regularization to NLG models for expressing constraints and desired properties related to a given task; and
3) the formal rules and structures that symbolic knowledge established can be used to improve the reasoning ability of NLG models.

In this paper, we stress the importance and necessity of NLG techniques to be guided by knowledge, so that human-like reasoning can be conveyed through language generation.
To this end, we propose ten goals for intelligent NLG systems to pursue ($\mathsection$ \ref{sec:goal}),
Next, we briefly review the achievement of NLG techniques with knowledge ($\mathsection$ \ref{sec:knowledge}), picturing the mutual enhancement of guiding NLG with knowledge and acquiring knowledge with NLG.
Then, we summarize how to make rational usage of knowledge to approach human-level reasoning with NLG techniques ($\mathsection$ \ref{sec:reasoning}), showing reasoning-guided NLG and how NLG can be used to chain up reasoning.
Finally, we conclude this paper and envision future directions and challenges in the pursuit of these goals ($\mathsection$ \ref{sec:conclusion}).

\section{Holy Grails in NLG}
\label{sec:goal}

\begin{figure}[t]
    \centering
    \includegraphics[width=7.5cm]{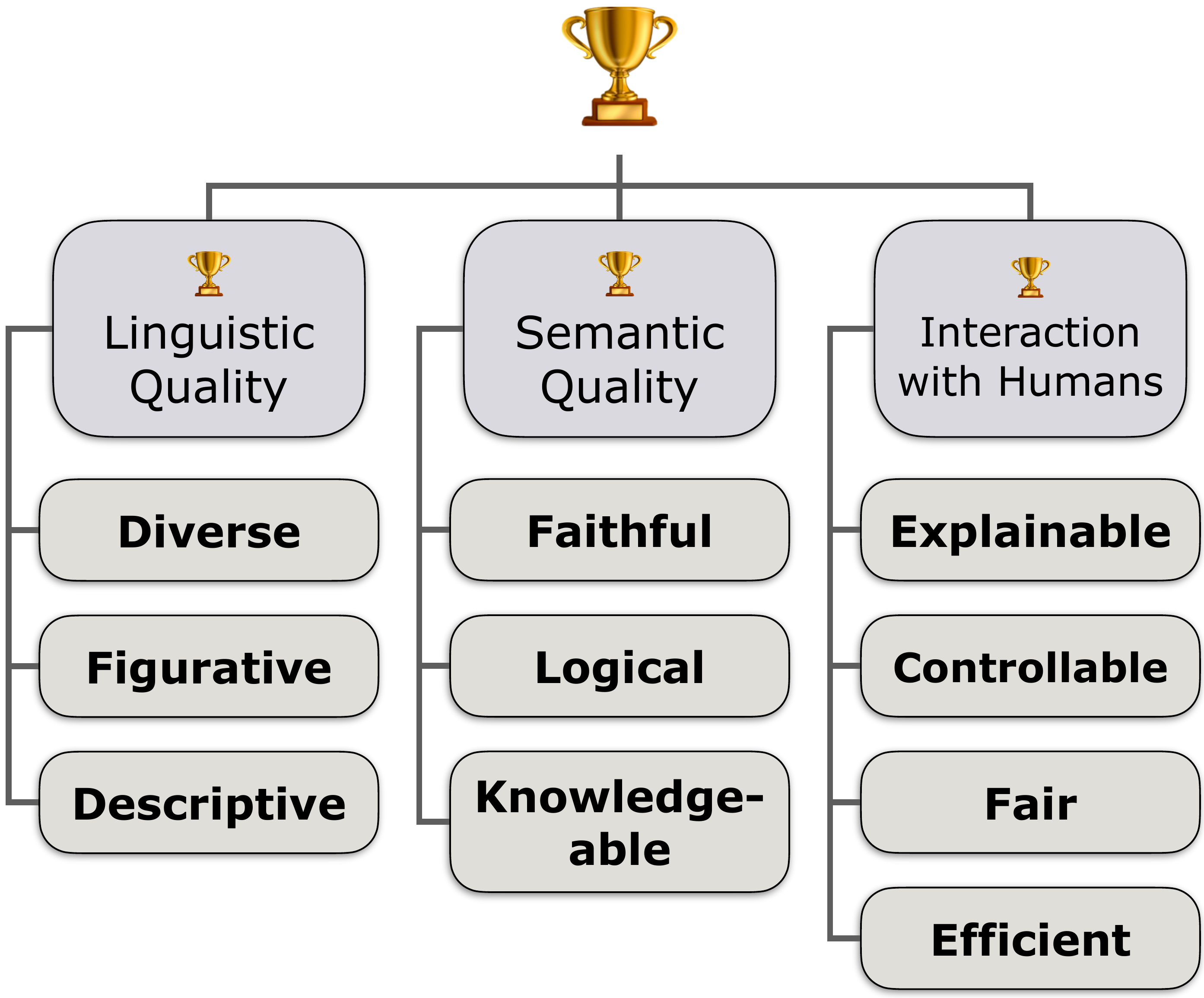}
    \caption{The taxonomy of the goals in NLG.}
    \label{fig:taxonomy}
\end{figure}

To begin with, we list ten most-desired goals (with room for more) for machine-generated text, in order to build an AI system that can freely interact and communicate with humans with language.
We categorize these goals into three classes based on the linguistic quality of generated text, conveying semantic information with generated text and the interaction with humans, as shown in Figure~\ref{fig:taxonomy}.
These goals are still quite challenging for modern NLG methods.
While the research about most of them is still in its infancy, some of these goals have been moderately explored by the community.
Since symbolic knowledge offers control of the process and outcome of generated text, many of research endeavors prove that symbolic knowledge plays an important role in steering NLG techniques towards these goals.

\paragraph{Diverse}

Given a source input, an ideal NLG model needs to be able to generate multiple and diverse outputs, where each of them needs to be equally valid.
For example
Diversified NLG is a practical goal, especially in the context of real-life applications such as dialogue generation~\cite{wu-etal-2020-diverse}, machine translation~\cite{shen2019mixture}, headline or query generation in e-commerce~\cite{10.1145/3488560.3498431} and text paraphrasing~\cite{10.1007/978-3-030-88480-2_28}.
Advanced NLG models should be able to generate diverse but informative text from the large sentence space, especially involving diverse but relevant background knowledge in the context.

\paragraph{Figurative}

The generative text should be figurative to build up emotional significance to the readers.
Figurative text typically includes idioms, sarcasm, simile, metaphor, etc.
All of them are great ingredients in interesting and creative writing tasks such as story generation or narratives.
For example, the sentence that ``he is drowning in a sea of grief'' expresses strong negative emotions to the reader, where \textit{grief}, with the metaphor of a \textit{sea}, vividly overwhelms \textit{him}.
To master the ability to write figurative text is non-trivial, and recent studies show that even strong modern language models still struggle at this objective~\cite{he-etal-2022-pre,liu-etal-2022-testing,chen-etal-2022-e}.
Nevertheless, research shows that such a problem can be alleviated with knowledge-enhanced models, which enrich the context and constituents of figurative text with acquired knowledge~\cite{chakrabarty-etal-2022-rocket}.

\paragraph{Descriptive}

Advanced NLG techniques need to be descriptive, that is, vivid and colorful as if something is being experienced by the readers.
Descriptive texts are rather common in books and novels, fascinated by image-like texts.
For example, to describe the sunset: ``\textit{the sunset filled the entire sky with the deep color of rubies, setting the clouds ablaze.}''\footnote{
https://rescuewriting.org/featured/writing-descriptive-text/.}
It is worth noting that, unlike previous desired properties that interact mostly with textual data, a system needs to integrate multi-modal knowledge and reasoning (e.g., image, video, etc.) to be descriptive with visual-specific features ~\cite{yin-ordonez-2017-obj2text,DBLP:conf/ai/Shi0Z21,shi-etal-2021-enhancing}.
However, the exact definition of descriptiveness in the context of the machine-generated text as well as its automatic evaluation are still great challenges.

\paragraph{Faithful} 

The generated text should be faithful to the input so that it correctly conveys and extends the input information without semantic violation.
Otherwise, the credibility of an NLG system could be undermined when hallucinated texts are generated, which could be dangerous sometimes.
There is a growing interest in enhancing and evaluating the (intrinsic or extrinsic) faithfulness of generated text in text summarization~\cite{cao2018faithful,maynez-etal-2020-faithfulness,ladhak-etal-2022-faithful}, dialogue systems~\cite{honovich-etal-2021-q2}, text simplification~\cite{devaraj-etal-2022-evaluating}, etc.
However, hallucinated texts are not always non-factual, because they may be faithful towards world knowledge~\cite{cao-etal-2022-hallucinated}.
Therefore, an NLG system should be faithful and can be verified by world knowledge~\cite{thorne-etal-2018-fever,Chen_Bao_Sun_Zhang_Chen_Zhou_Xiao_Li_2022}, except for applications such as fictional story generation.

\paragraph{Logical} 

Logical reasoning is an essential part of human thinking and language; therefore, the generated text needs to be logically consistent and self-contained.
Different from faithfulness which focuses on information consistency, logical consistency poses a challenge over the discourse of produced language of an NLG system~\cite{betz-etal-2021-critical,chen-etal-2020-logical,shi-etal-2021-refine-imitate,shu-etal-2021-logic,pi2022logigan}.
However, the training objectives of current prevailing language modeling (e.g., masked language modeling and causal language modeling) prioritize recovering given text, which does not guarantee the logical reasoning ability to be effectively captured by models.

\paragraph{Knowledgeable}

An NLG model needs to be knowledgeable to generate text that is rich in knowledge~\cite{yu2022survey}.
When engaging in a conversation, current NLG models are known to be dependent on plain but safe responses.
A knowledgeable NLG system, in contrast, should be able to actively initiate responses grounded with the background knowledge that is either retrieved~\cite{NEURIPS2020_6b493230} or inherent~\cite{liu-etal-2022-multi} in the system itself.
Also, it is worth exploring whether current NLG models are using knowledge of their own or just relying on statistical patterns learned from (pre-)training~\cite{cao-etal-2021-knowledgeable}.

\paragraph{Explainable}
An NLG system should be explainable for the trust of human users.
Since language is the most natural tool for communication, the reasoning and decisions of a model should be explanatory through generated language~\cite{reiter2019natural}. 
Existing work usually focuses on generating natural language explanations~\cite{wiegreffe-marasovic-2021-review} as a task, while how to develop universal explanation generators~\cite{ye2022unreliability} is still challenging.
More importantly, unlike natural language understanding models (NLU)~\cite{lime,Schulz2020Restricting,10.5555/3305890.3306024}, the explainability of NLG systems is severely underexplored.

\paragraph{Controllable}

Controllable NLG aims to generate with desired attributes by users, which greatly broadens its applications.
Current commonly used attributes focus on discriminative attributes such as sentiment and style, lexical constraints, and various properties of language such as lengths or complexity~\cite{NEURIPS2021_d0f5edad,zhang2022survey,garbacea2022constrained}.
Moreover, there are still many interesting open questions for building an ideal NLG system, for example, more diverse applications of controlling factors, satisfying multiple types of constrained attributes (perhaps a mixture of them), and being controlled in a few-shot or even zero-shot manner.
It is worth noting that recent studies in large language models~\cite{NEURIPS2020_1457c0d6} shed some light on these research problems.

\paragraph{Fair} 

The generated text should be fair and contain no bias whatsoever.
Methods to mitigate biases have been proposed w.r.t. gender, race, nationality, age, disability, religion, etc.
\cite{liu-etal-2020-gender,xia-etal-2020-demoting,gupta-etal-2022-mitigating}.
Since text generation models have been widely applied in the age of the Internet, their outstanding performance could blind service providers so as to omit the negative social impacts that an unfair NLG system causes.
However, since modern NLG systems are usually pre-trained on colossal unannotated corpus~\cite{radford2019language,raffel2020exploring,lewis-etal-2020-bart,NEURIPS2020_1457c0d6}, the biases within them are still rather difficult to eliminate.

\paragraph{Efficient} 

An NLG system should be efficient to be deployed on high-demanding scenarios such as mobile devices or online industrial applications.
Recent work on non-autoregressive generation (NAG)~\cite{gu2018nonautoregressive,xiao2022survey}, which generates tokens in parallel, has shown promising potential in efficiency.
This enables NAG to achieve over $10\times$ speedup over the autoregressive counterpart (AG) that generates in a left-to-right manner~\cite{vaswani2017attention}.
However, NAG is still haunted by the multi-modality problem~\cite{Zhou2020Understanding}, its generality to tasks other than machine translation~\cite{qian-etal-2021-glancing}, and interaction with external knowledge~\cite{zeng-etal-2022-neighbors}.

\section{Knowledge in Language Generation}
\label{sec:knowledge}

\begin{figure}[tp]
    \centering
    \includegraphics[width=7.5cm]{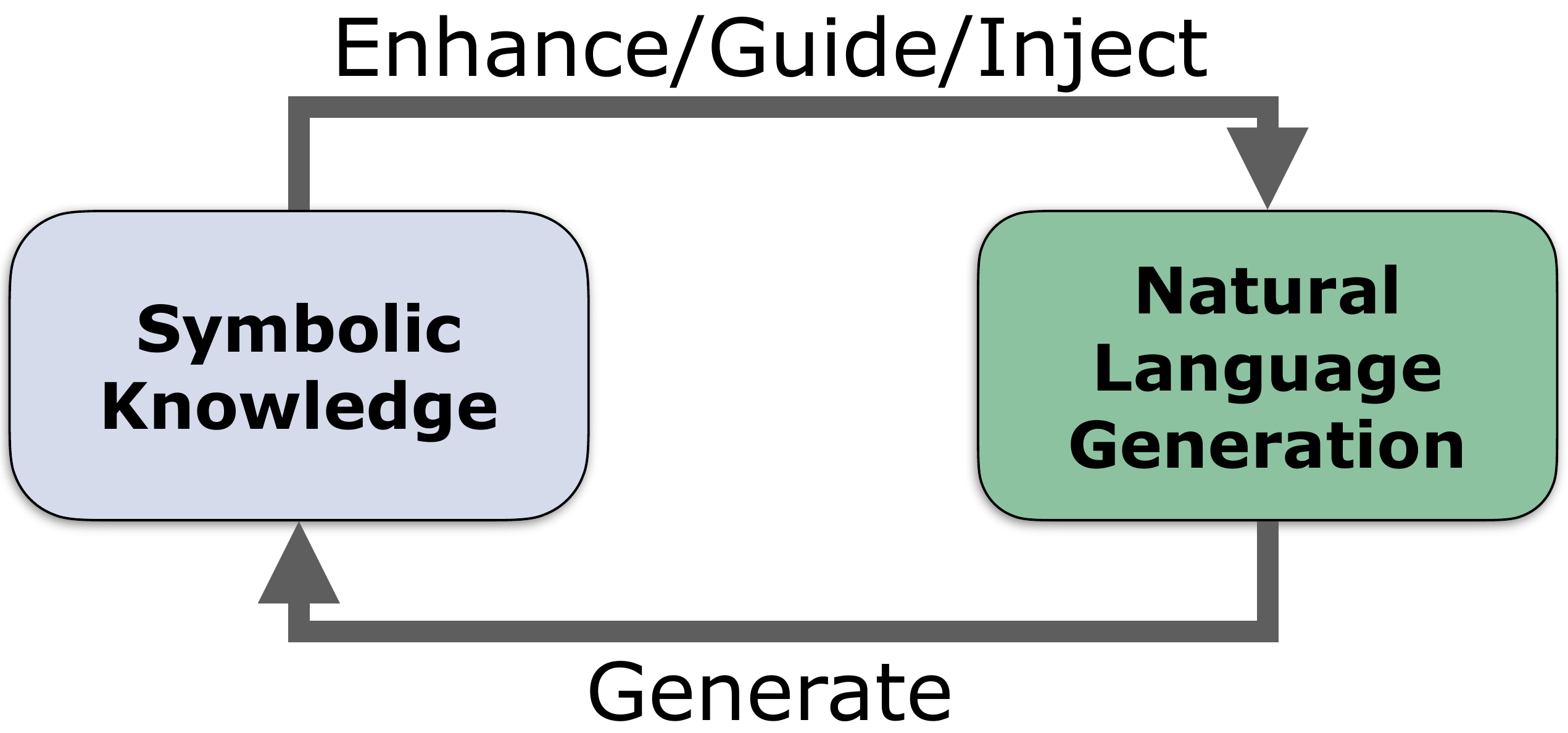}
    \caption{Relations between knowledge and NLG, where knowledge can enhance, guide and be injected into NLG systems, and NLG systems can be utilized as tools to generate new knowledge.}
    \label{fig:knowledge-nlg}
\end{figure}

In this section, we briefly review some representative work in knowledge-guided NLG algorithms and knowledge acquisition with NLG models, where NLG systems also show unique advantages in the latter.
Figure~\ref{fig:knowledge-nlg} shows the relation between knowledge-guided NLG ($\mathsection$~\ref{subsec:knowledge-guided-nlg}) and knowledge acquisition with NLG ($\mathsection$~\ref{subsec:generative_acquisition}), where the two are mutually beneficial to each other.

\subsection{Knowledge-guided NLG: Tricks of the Trade}
\label{subsec:knowledge-guided-nlg}

NLG systems can be guided by multiple sources and types of knowledge to be logical, diverse, faithful, controllable, etc.
In the following text, we list the knowledge sources that are commonly used, and showcase the common tricks of current knowledge-guided NLG algorithms, presenting a brief but as comprehensive review as possible.

\paragraph{Knowledge Sources}

According to \citet{yu2022survey}, knowledge sources used in NLG systems can be categorized into internal and external knowledge, where: 
1) internal knowledge is created within the given text, e.g., keywords, topics, linguistic features, etc., 
and 2) external knowledge exists outside the input of the NLG system, e.g., knowledge bases/graphs, grounded unstructured text.
We add that knowledge mined from the corpus is also important to guide NLG systems, such as concepts, rules, and patterns~\cite{chen-etal-2019-ensuring}.
Notably, NLG systems form an interesting dual learning loop where knowledge can be acquired by NLG systems ($\mathsection$ \ref{subsec:generative_acquisition}) and used in NLG systems~\cite{sun-etal-2018-logician,cao-etal-2020-unsupervised-dual}.
Next, we will detail some common practices of NLG models that involve knowledge.

\paragraph{Knowledge Incorporation in Model Architectures}
Modifying model architectures to incorporate knowledge is one of the most commonly used approaches in knowledge-guided NLG~\cite{yang2021survey,yu2022survey}.
Typical methods include using attention mechanism~\cite{DBLP:journals/corr/BahdanauCB14} (and its variants, e.g., copying mechanism,~\citealt{gu-etal-2016-incorporating}) to attend additional knowledge sources, encoding knowledge with specific encoders (e.g., graph neural networks,~\citealt{kipf2016semi}), adding specific layers in the neural network for knowledge storage (e.g., adapter layers,~\citealt{guo2020incorporating}), etc.
In this line of work, a model trains to utilize the encoded knowledge for better NLG.
However, such methods usually require the design of an ad hoc model architecture, which does not generalize well to new tasks, especially in the age of large PLMs~\cite{NEURIPS2020_1457c0d6}.

\paragraph{Learning with Knowledge}

The most common method of knowledge-guided NLG is through supervised learning, where the guidance of knowledge can be well integrated into generation~\cite{yu2022survey}.
A simple way to do this is to append the acquired knowledge to the input as an additional context.
This has been shown to be effective, especially when knowledge acquisition (e.g., retrieval) is jointly trained with generation~\cite{pmlr-v119-guu20a,NEURIPS2020_6b493230}.
Another way is to guide learning objectives with knowledge.
Studies in this direction are carried out by designing knowledge into one of the training targets~\cite{cao-etal-2020-unsupervised-dual} or pre-training~\cite{guan-etal-2020-knowledge}, or by introducing additional knowledge-related objective functions to regularize training~\cite{hu2018deep}.
Also, reward design in a reinforcement learning (RL) framework is a flexible way of integrating knowledge into training objectives for NLG models.
For an example of improving the consistency between source and target, \citet{li2018ensure} inject the entailment knowledge into NLG with an entailment model, and \citet{huang-etal-2020-knowledge} design a semantic cloze reward for faithful summarization.

\paragraph{Knowledge-Constrained Decoding}

Finally, we would like to emphasize knowledge incorporation during decoding.
Knowledge injection through model architecture modification and new learning objectives, despite being effective, requires (re-)training.
This usually causes inconvenience for deployment, especially with large PLMs.
For this reason, constraining text decoding in inference time with knowledge has become a promising research topic~\cite{zhang2022survey,garbacea2022constrained}.
In this line of work, we briefly introduce the three most exemplary knowledge constraint types in NLG systems: 
1) lexical knowledge constraints, such as keywords that must appear in the text, 
2) discriminative constraints, such as the desired attributes of the generated text, and 
3) structured knowledge constraints, such as the world knowledge that the generated text must be consistent with.

Imposing \textit{lexical constraints} is challenging for NLG systems because the prevailing ones usually generate in an autoregressive manner, which is difficult to know when and where to keep the constraints.
Popular solutions to this problem mainly focus on adding constraints during beam search~\cite{hokamp-liu-2017-lexically,post-vilar-2018-fast,lu-etal-2021-neurologic} at the cost of computational complexity.
Recent work~\cite{susanto-etal-2020-lexically,10.1162/tacl_a_00368} also explores iterative lexically constrained decoding for non-autoregressive models, achieving significant speedups.

\textit{Discriminative constraints} usually involve prior attribute models to control the generated text in terms of attributes such as topics (sport, finance, medical), sentiments (positive, negative), etc.
Exemplary work is PPLM ~\cite{Dathathri2020Plug}, which controls the attribute of generated text without changing the parameters of the backbone PLM (e.g., GPT-2,~\citealt{radford2019language}).
During sampling, PPLMs back-propagates the gradient from the attribute model to the hidden states of the PLM so that the generation can be steered by the attribute model.
Under such a framework, it is easy to guide generation with prior knowledge.
It is worth noting that work on text sampling is also a great framework to effectively incorporate the above-mentioned two types of constraints.
In this framework, the desired properties are designed as objective functions, guiding the sampling process, which is usually instantiated with Markov chain Monte Carlo methods~\cite{Miao_Zhou_Mou_Yan_Li_2019,zhang-etal-2020-language-generation,qin2022cold}.
Similar to PPLM, such methods ensure the generated text is constrained to the desired objectives without training.

Recent studies have also started to pay attention to the hallucination problem of NLG systems (represented by PLMs) from the perspective of constrained decoding with \textit{structured knowledge constraints}, such as knowledge graphs.
For example, \citet{liu2022knowledge} retrieve from knowledge graphs related to neighboring entities within source input.
They are used to modify and constrain the distributions over the vocabulary during sampling, where tokens within knowledge graphs are rewarded.
In this way, the generated text would be rich in semantic knowledge and faithful to world knowledge, making it an interesting and effective solution to the hallucination problem that is common in NLG.


\subsection{Knowledge Acquisition with NLG}
\label{subsec:generative_acquisition}
Knowledge acquisition based on NLG systems aims to generate knowledge from the input text.
Compared with other paradigms of knowledge acquisition, automatic and generalizability to unseen knowledge are one of the most desired features of generative methods.
In the following text, we first discuss current knowledge acquisition paradigms and show the advantages of generative methods in certain scenarios.

\paragraph{Paradigms of Knowledge Acquisition}

The knowledge acquisition methods can be divided into three categories: 1) crowd-sourcing, 2) extractive and 3) generative methods.
The \textit{crowd-sourcing} methods~\cite{10.1145/219717.219748} invite some experts to label the knowledge in the corpus, which can acquire accurate knowledge but only have a small size due to costly annotation.
The \textit{extractive} methods~\cite{zhou2022survey} adopt language models to extract the knowledge explicitly mentioned in the input text automatically but ignore some symbolic and neural commonsense knowledge.
Compared with them, generative knowledge acquisition methods enjoy the advantages of the ability to generate explicit and implicit knowledge, which we will soon discuss.

\paragraph{Meta Knowledge Generation}
Meta knowledge is knowledge about knowledge~\cite{Davis1977MetaLevelKO}. 
As a fundamental conceptual instrument in knowledge-based domains, meta-knowledge can greatly improve the performance of downstream tasks, such as text classification~\cite{chen2019deep}, question answering~\cite{garderes-etal-2020-conceptbert} and story generation~\cite{chen2021commonsense}. 
However, meta-knowledge is hardly explicitly mentioned in the corpus, and thus it is difficult to directly extract meta-knowledge from the text. 
Alternatively, since NLG systems can generate information never mentioned in the text and has a strong generalization ability of unseen knowledge, recent work proposes to adopt generative methods to acquire meta knowledge, such as concepts~\cite{chen-etal-2019-ensuring,9409694} and rules~\cite{10.1016/j.knosys.2022.108371}. 

\paragraph{Generative Knowledge Retrieval}
Information retrieval aims to retrieve meaningful information from large Knowledge Bases (KB) given a textual input.
Compared to extractive methods, generative information retrieval~\cite{cao2021autoregressive,10.1162/tacl_a_00460,rossiello2021generative} can directly capture the relation between context and target information and reduce the memory footprint without negative data down-sampling.

\paragraph{Generative Knowledge Extraction}
Information extraction aims to obtain information from semi-structured and unstructured text, which suffers heterogeneous structures and domain-specific schemas~\cite{lu-etal-2022-unified}.
The generative information extraction can end-to-end generate targeted structures directly.
For example, \citet{huguet-cabot-navigli-2021-rebel-relation} adopt autoregressive models to achieve the extraction of end-to-end relations.
\citet{lu-etal-2021-text2event} propose a sequence-to-structure generation method to directly extract events from the text.
\citet{huang-etal-2021-document} formulate entity-based extraction as a template generation task to allow the generative framework to effectively capture cross-entity dependencies.

\paragraph{Knowledge Generation}

NLG models can also be used for the completion of knowledge bases, including commonsense knowledge graphs and rules.
Instead of extracting semi-structured and unstructured text into knowledge, some work~\cite{bosselut-etal-2019-comet,Hwang2021COMETATOMIC2O} feeds large-scale language models with a massive corpus to obtain knowledge models, which can adapt their learned representations to knowledge generation and automatically construct KBs. 
Furthermore, the parameters of PLMs are shown to store vast amounts of linguistic knowledge~\cite{47786}. 
A line of work further regards PLMs as knowledge bases and distills semantic knowledge from these models~\cite{petroni-etal-2019-language,alkhamissi2022review}.

\section{Reasoning in Language Generation}
\label{sec:reasoning}

We echo the argument that good reasoning should be right for the right reasons.
Therefore, it is also crucial for an NLG model to make \textit{rational} usage of knowledge to approach human-like reasoning skills.
This section signifies the importance of reasoning in NLG, where we sketch two lines of research: 1) reasoning-guided NLG methods ($\mathsection$~\ref{subsec:reasoning-guided-nlg}) and 2) NLG for the purpose of reasoning ($\mathsection$~\ref{subsec:reasoning-by-nlg}).

\subsection{Reasoning-guided NLG}
\label{subsec:reasoning-guided-nlg}
In contrast to general knowledge-guided NLG, reasoning-guided NLG systems make more rational and explainable usage of the knowledge (in wide forms).
Since reasoning is highly correlated with knowledge ($\mathsection$ \ref{sec:knowledge}), we will discuss reasoning-guided NLG by highlighting the topics of graph reasoning and generative reasoning tasks in the following paragraphs.

\paragraph{Graph Reasoning}

Graph reasoning is one of the most commonly used techniques to guide NLP systems towards more multihop, controllable, and explainable reasoning.
Therefore, we extend what has been discussed in $\mathsection$ \ref{subsec:knowledge-guided-nlg} in this paragraph for introducing graph reasoning-guided NLG.
Graph reasoning implementations are usually built on retrieved subgraphs of external knowledge graphs~\cite{Liu_Wan_He_Peng_Yu_2021} or internal graphs parsed from the input~\cite{wang-etal-2020-heterogeneous}.
Most of them adopt graph embeddings~\cite{NIPS2013_1cecc7a7} and graph neural networks~\cite{kipf2016semi,velickovic2018graph} to propagate information throughout the graph.
These properties of graphs enable the multi-hop reasoning ability of NLG models for long-range text~\cite{ji-etal-2020-language} and generating emerging concepts or topics guided by the graph~\cite{wang-etal-2019-paperrobot,zhang-etal-2020-grounded}.
Due to the symbolic structure of graphs, heuristics and prior knowledge can be easily incorporated into graph construction, e.g., building graphs with various parsing tools ~\cite{gardner2017allennlp} such as semantic role labeling or dependency parsing or guiding graph propagation with more information, such as popularity knowledge~\cite{10.1145/3488560.3498431}.
Also, such methods are explainable and easy to debug, since the weights on the graph nodes and edges greatly facilitate post hoc manual examination.

\paragraph{Reasoning Tasks}

One of the most direct ways to enable NLG models to reason is to design various reasoning tasks.
During the solving of these tasks, researchers can develop and test their models w.r.t. corresponding reasoning skills, which makes it an important direction to guide AI models.
Over the years, the community has accumulated many datasets of tasks that test various facets of machine reasoning in the form of text generation.
Most of them are built from the point of view of human cognition, including tasks about logical reasoning~\cite{dalvi-etal-2021-explaining}, abductive reasoning~\cite{bhagavatula2020abductive}, counterfactual reasoning~\cite{qin-etal-2019-counterfactual}, generative commonsense reasoning~\cite{lin-etal-2020-commongen}, social reasoning~\cite{sap-etal-2019-social}, physical reasoning~\cite{Bisk_Zellers_Le_bras_Gao_Choi_2020}, temporal reasoning~\cite{zhou-etal-2019-going}, etc. \footnote{Note that some of these reasoning tasks~\cite{sap-etal-2019-social,Bisk_Zellers_Le_bras_Gao_Choi_2020,zhou-etal-2019-going} take the form of question answering but can be solved in a generative manner~\cite{2020unifiedqa}.}
Not limited to these tasks, datasets on explanation generation~\cite{wiegreffe-marasovic-2021-review} are also good sources to evaluate and improve the reasoning ability of NLG systems, including the explanations for natural language inference~\cite{NEURIPS2018_4c7a167b} commonsense reasoning~\cite{rajani-etal-2019-explain}, analogical reasoning~\cite{chen-etal-2022-e}, causal reasoning~\cite{du-etal-2022-e}, multimodal reasoning~\cite{li2018vqae}, etc.





\subsection{Reasoning by NLG}
\label{subsec:reasoning-by-nlg}

Human language is a good vehicle for reasoning.
Thus, the generation of language naturally resembles the way humans think and reason.
With the success of PLMs, there is a growing interest in the AI community to use NLG models to generate a chain of reasoning for problem-solving.

\paragraph{Generative Reasoning}

Generative reasoning aims to generate intermediate reasons with NLG models for better problem solving.
Such intermediate reasons take many forms, including deduction and abduction reasons~\cite{tafjord-etal-2021-proofwriter},  explanations~\cite{jhamtani-clark-2020-learning}, or decomposed subtasks of a complex one~\cite{khot-etal-2021-text}. 
Some work~\cite{shwartz-etal-2020-unsupervised,betz2021thinking} even show that expanding the context of the input by generating more information would also help solve reasoning tasks.
Moreover, \citet{ijcai2020-0537} find that transformer-based PLMs are effective soft reasoners on a toy deduction dataset, which consists of collections of text verbalized from artificial if-then rules and facts.
Other studies~\cite{bostrom-etal-2021-flexible,betz-etal-2021-critical} corroborate this discovery and show that training generative models with artificial textual data that verbalize rule-based reasoning helps downstream logical reasoning tasks.
We remark that generative reasoning provides a new perspective of breaking the black-box prediction of neural networks, which demonstrates the potential of achieving reasoning with NLG systems.

\paragraph{Reasoning with Large Language Models}

Entering the era of large language models (LLMs) such as GPT-3~\cite{NEURIPS2020_1457c0d6,ouyang2022training} and PaLM~\cite{chowdhery2022palm}, there is a recent growing interest in exploring few-/zero-shot reasoning skills of these LLMs
Since fine-tuning such tremendous language models is hardly possible, current work adopts prompt-based in-context learning methods~\cite{NEURIPS2020_1457c0d6,lu-etal-2022-fantastically} to achieve few-/zero-shot learning with LLMs, where instructions and examples are demonstrated within the input prompts.

Similar to the above discussion of generative reasoning, this line of work aims to guide the language models to explicitly generate the intermediate thinking steps (or reasons) during reasoning.
A representative work among them is the Chain-of-Thought prompting~\cite{wei2022chain}, where the intermediate thinking process is verbalized and integrated into the demonstrations.
In this way, complex reasoning can be decomposed into multiple steps reflected by language, which is analogous to how humans solve complex tasks.
Such methods achieve much better performance on a variety of reasoning tasks compared with normal prompting and even surpass fine-tuned methods in some cases.
Moreover, the prompting strategy can be further refined~\cite{wang2022self,zhou2022least,creswell2022selection}, leading to generally better results.
LLMs prompted with and asked to generate step-by-step reasoning chains also exhibit certain quantitative reasoning abilities such as solving math word problems~\cite{wei2022chain}, where the LLMs are not specifically trained on such tasks.
However, reasoning abilities for LLMs are shown to be emergent~\cite{wei2022chain,wei2022emergent}, i.e., effective only for really large language models (over 100 billion parameters).
How to enable smaller language models with few-shot reasoning skills is still an open question.

\section{Conclusion}
\label{sec:conclusion}

In this work, we envision the ten most desired goals of an intelligent natural language generation system.
In pursuit of these goals, the guidance of knowledge and reasoning plays a significant role in modern NLG models.
We have revisited the achievements with knowledge in NLG w.r.t. knowledge-guided NLG and generative knowledge acquisition.
We particularly highlight knowledge-constrained decoding for its wide application potential, where knowledge constraints can be incorporated into NLG models (especially large-scale ones) in a plug-and-play manner.
We have also discussed current work on reasoning in NLG, which essentially makes rational usage of knowledge and datasets for various reasoning tasks.
Moreover, NLG can verbalize intermediate thinking processes to facilitate complex reasoning, enabling few-/zero-shot reasoning abilities for large language models.
However, current research is still far from realizing these goals, which we outline for future research. 
We hope that this survey report can provide newcomers with a good entry point into the exciting area of knowledge-guided and reasoning-intensive NLG.

\bibliography{custom}
\bibliographystyle{acl_natbib}

\end{document}